\documentclass[fleqn,10pt]{wlscirep}
\usepackage[utf8]{inputenc}
\usepackage[T1]{fontenc}
\usepackage{graphicx}

\usepackage{xargs}
\usepackage[textsize=scriptsize]{todonotes}\reversemarginpar

\newcommandx{\jeffNote}[2][1=]{\todo[inline,linecolor=black,backgroundcolor=orange!25,bordercolor=orange,#1]{#2 ---Jeff}}
\newcommandx{\andreasNote}[2][1=]{\todo[inline,linecolor=black,backgroundcolor=red!25,bordercolor=red,#1]{#2 ---Andreas}}
\newcommandx{\toddNote}[2][1=]{\todo[inline,linecolor=black,backgroundcolor=blue!25,bordercolor=blue,#1]{#2 ---Todd}}
\newcommandx{\salarNote}[2][1=]{\todo[inline,linecolor=black,backgroundcolor=green!25,bordercolor=green,#1]{#2 ---Salar}}
\newcommandx{\jimNote}[2][1=]{\todo[inline,linecolor=black,backgroundcolor=purple!25,bordercolor=purple,#1]{#2 ---Jim}}
\newcommandx{\williamNote}[2][1=]{\todo[inline,linecolor=black,backgroundcolor=yellow!25,bordercolor=yellow,#1]{#2 ---William}}

\title{Making BREAD: Biomimetic strategies for Artificial Intelligence Now and in the Future}

\author[1*]{Jeffrey L. Krichmar}

\author[2]{William Severa}
\author[3]{Salar M. Khan}
\author[4+]{James L. Olds}
\affil[1]{University of California, Irvine, Department of Cognitive Sciences, Department of Computer Science, Irvine, CA, 92697-5100, USA}

\affil[2]{Sandia National Laboratories, Data-Driven and Neural Computing, Albuquerque, NM, 87123, USA}
\affil[3]{George Mason University, Schar School, Arlington VA, 22201 USA}
\affil[4]{George Mason University, Schar School, Arlington VA, 22201, USA}

\affil[*]{jkrichma@uci.edu}
\affil[+]{jolds@gmu.edu}


\begin{abstract}
The Artificial Intelligence (AI) revolution foretold of during the 1960s is well underway in the second decade of the 21st century. Its period of phenomenal growth likely lies ahead. AI-operated machines and technologies will extend the reach of Homo sapiens far beyond the biological constraints imposed by evolution: outwards further into deep space as well as inwards into the nano-world of DNA sequences and relevant medical applications. And yet, we believe, there are crucial lessons that biology can offer that will enable a prosperous future for AI. For machines in general, and for AI’s especially, operating over extended periods or in extreme environments will require energy usage orders of magnitudes more efficient than exists today. In many operational environments, energy sources will be constrained. The AI’s design and function may be dependent upon the type of energy source, as well as its availability and accessibility. Any plans for AI devices operating in a challenging environment must begin with the question of how they are powered, where fuel is located, how energy is stored and made available to the machine, and how long the machine can operate on specific energy units. While one of the key advantages of AI use is to reduce the dimensionality of a complex problem for which the required computation is thermodynamically expensive and therefore energy-constrained, the fact remains that some energy is required for functionality.  Hence, the materials and technologies that provide the needed energy represent a critical challenge towards future use-scenarios of AI and should be integrated into their design. Here we make four recommendations for stakeholders and especially decision makers to facilitate a successful trajectory for this technology. First, that scientific societies and governments coordinate Biomimetic Research for Energy-efficient, AI Designs (BREAD)— a multinational initiative and a funding strategy for investments in the future integrated design of energetics into AI. Second, that biomimetic energetic solutions be central to design consideration for future AI. Third, that a pre-competitive space be organized between stakeholder partners and fourth, that a trainee pipeline be established to ensure the human capital required for success in this area.
\end{abstract}
\begin{document}

\flushbottom
\maketitle
%
%
\thispagestyle{empty}

\section*{Artificial Intelligence's Energy Requirements}

The last few years have seen an explosion of Artificial Intelligence (AI) and Machine Learning (ML) breakthroughs. What were once AI solutions to small, toy problems have now become human level complex problem-solving. These solutions have moved out of research labs and into commercial applications. However, most AI and ML algorithms for these complex problems are implemented in large data centers housing power hungry clusters of computers and Graphical Processing Units (GPUs). In contrast, natural, biological intelligence is power efficient and self-sufficient. In this article, we argue for Biomimetic Research for Energy-efficient, AI Designs (BREAD) as AI moves toward edge computing in remote environments far away from conventional energy sources, and as energy consumption becomes increasingly expensive.

\begin{figure}[!ht]
\centering
\includegraphics[width=1\linewidth]{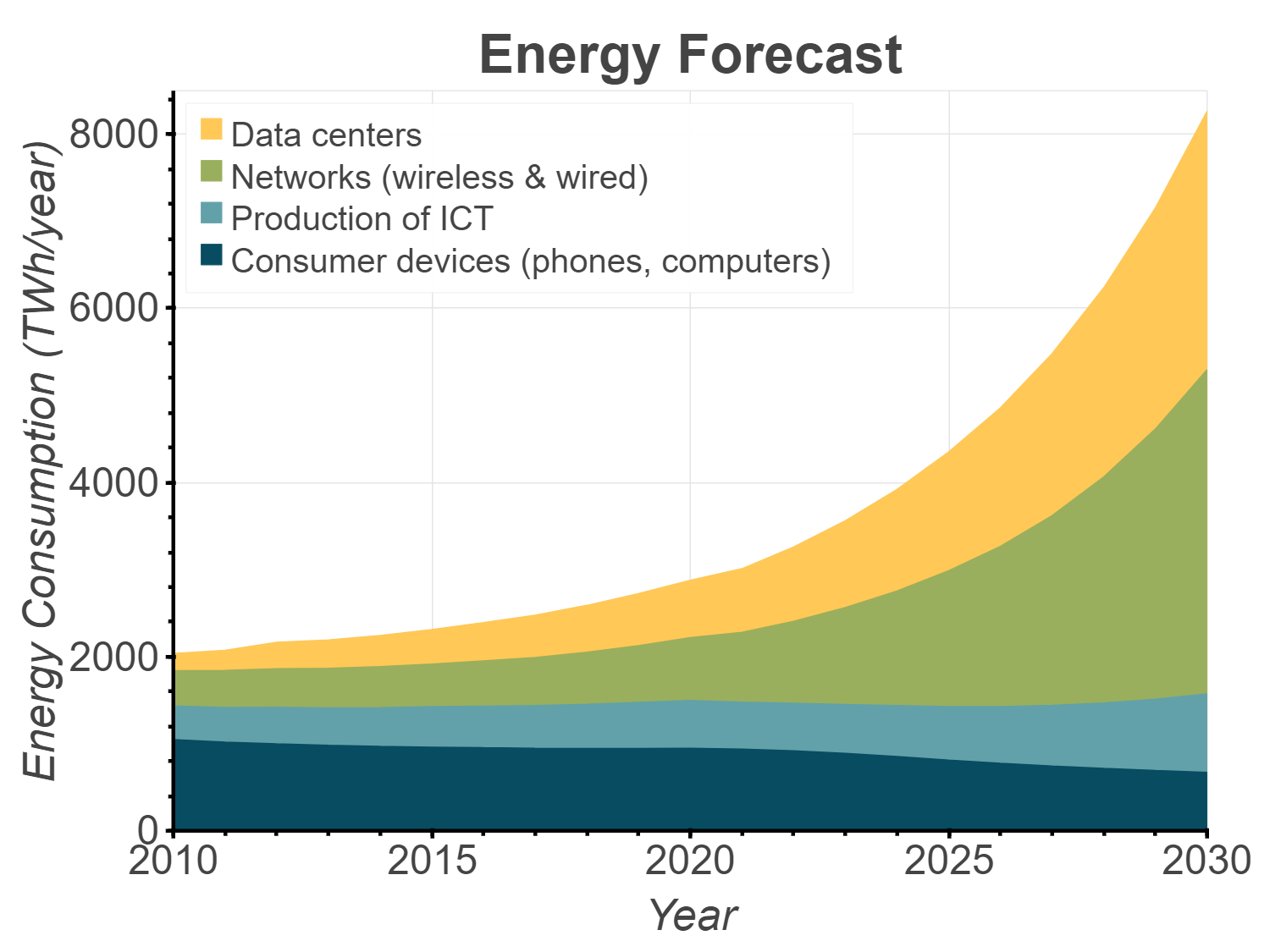}
\caption{Energy demand will increase: The graph shows that the total ICT electricity demand is rising in the 2020s. Data centers will likely constitute a bulk of this demand. (Chart adapted from \cite{RN52})}
\label{fig:forecast}
\end{figure}

With the growth of the Internet, data traffic (traffic to and from data centers) is escalating exponentially, crossing a zettabyte (1.1 ZB) in 2017 \cite{RN52}. Figure \ref{fig:forecast} shows this trend. Currently, data centers consume an assessed 200 terawatt hours (TWh) each year equivalent to 1 percent of global electricity demand \cite{RN33}.  A 2017 International Energy Agency (IEA) report notes that even with the ongoing explosion of Internet traffic in data centers, efficiency gains will result in increased electricity demand only by 3 percent \cite{RN53}.

\subsection*{Current Solutions to AI's Energy Requirements}
\begin{figure}[!ht]
\centering
\includegraphics[width=1\linewidth]{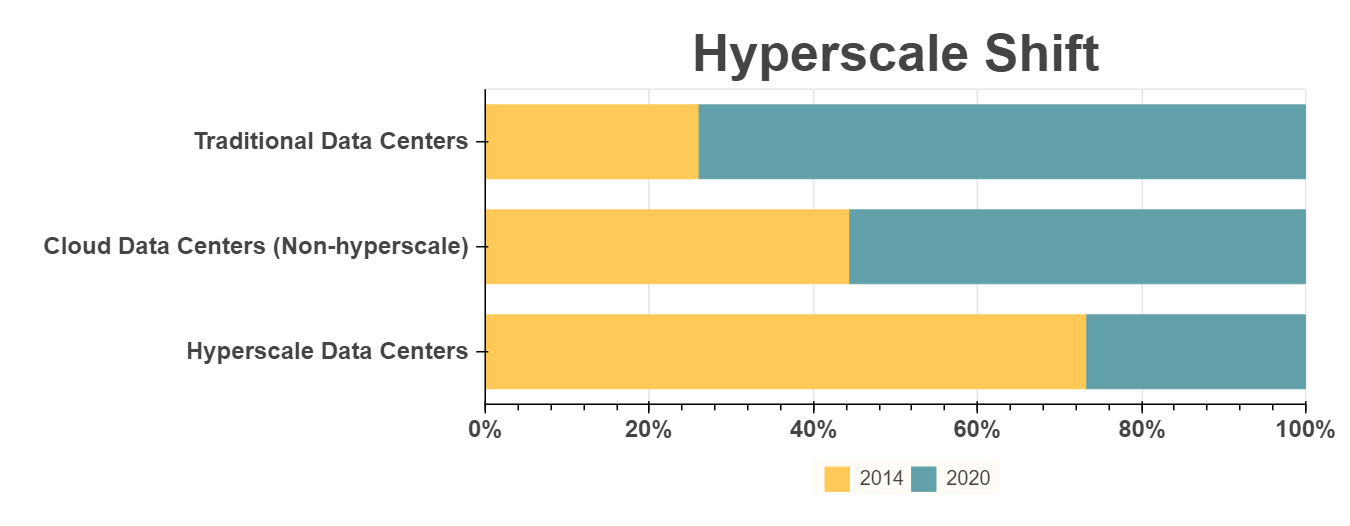}
\caption{As smaller and inefficient data centers shut down, efficient hyperscale data centers are anticipated to make half of the data-center electricity demand by 2020. Image adapted from \cite{RN53}.}
\label{fig:hyperscale}
\end{figure}

The rise of highly efficient data factories, known as hyperscale facilities, use an organized uniform computing architecture that scales up to hundreds of thousands of servers (see Figure \ref{fig:hyperscale}). While these hyperscale centers can be optimized for high computing efficiency, there are limits to growth due to a variety of constraints that also affect other electrical grid consumers. However, the shift to hyperscale facilities is a current trend, and if 80 percent of servers in US conventional data centers were moved over to hyperscale facilities, energy consumption would drop by 25 percent, according to Lawrence Berkeley National Laboratory report, 2016.   

On average, one server in a hyperscale center can replace $3.75$ servers in a conventional center. Hyperscale centers have a lower Power Usage Efficiency (PUE) than conventional data centers;  conventional data centers typically have a PUE of $2.0$, while for hyperscale centers PUE is approximately $1.2$~\cite{RN33,RN53}. One way the hyperscale centers have cut down their PUE is through efficiencies in cooling. By locating in cooler areas,  the data centers can ingest the cool air outside with positive results. Another solution is employing warm water cooling loops, a solution tuned for temperate and warm climates. An innovative solution to address the energy constraints of AI systems is to employ an AI-powered cloud-based control recommendation system.  For example, Google employs a cloud-based AI to collect information about the data center cooling system from thousands of physical sensors prior to feeding this information into deep neural networks. The networks then compute predictions for how different combinations of possible activities will affect future energy consumption \cite{RN54}.

Although hyperscale centers and smart cooling strategies can lower energy consumption, these solutions do not address applications where AI is operating at the edge or when AI is deployed in extreme conditions far away from convenient power supplies. We believe that this is where future AI systems are headed.
  
Our view is that there is a pressing need to address the energy issue as it applies to the future of AI and ML. While there is a growing research effort towards developing efficient machine learning methods for embedded and neuromorphic processors~\cite{howard2017mobilenets, severa2018whetstone,hunsberger2016training,esser2015backpropagation,rastegari2016xnor}, we recognize that these methods do not address the full needs of future applications, despite offering compelling first steps. Generally, current methods modify existing techniques rather than develop de novo algorithms. We emphasize how biological evolution has addressed this problem, and survey biologically inspired algorithms and computers. Furthermore, we discuss how the potential of future energy sources and computing architectures meet these power requirements. Finally, we recommend a set of policies to implement a coherent US government-wide investment strategy to ensure that future AI and ML are energetically practicable.

\section*{Current State}

\subsection*{Edge computing in Remote and Hostile Environments}

Trends in many human-built systems point to directions where sensing, processing, and actuation is situated on distributed platforms. The emerging Internet-of-Things (IoT) are cyber technologies \cite{RN58}, hardware and software, that interact with physical components in environments which may or may not be populated by humans. IoT devices are often thought of as the “edge” of a large, sophisticated cloud processing infrastructure. Processing data at the “edge,” reduces system latency by removing the delays in the aggregation tiers of the information technology infrastructure \cite{RN55,RN56,RN57}. In addition to minimizing latency, edge processing increases system security and mitigates privacy concerns when processing data in the cloud. Finally, in cases where the data path between the edge and user is very long, such computation can, by feature extraction, reduce the dimensionality and hence expense sending information. However, edge processing may be far away from power sources and may need to operate without intervention over long time periods.

Space and deep ocean exploration, which will most likely require AI and ML solutions, represent environments that are inherently hostile to the circuitry that sub-serves current AI and ML technologies. Gamma-ray bursts, solar weather, micrometeorite impacts are all exemplars of how space environments from Low Earth Orbit (LEO) through Deep Space are challenging. Planetary missions such as NASA's Curiosity Mars rover have revealed additional challenges from weather (such as sandstorms) that have put missions at risk. Radioisotope thermal generation power was added to the Curiosity Mars rover design to combat the vulnerabilities to solar energy systems on previous missions. While Curiosity's computational systems do not constitute true AI, their power demands are of a similar order of magnitude. Limitations on the availability of radioisotopes~\cite{aebersold1949production} combined with safety concerns~\cite{1948availability, al2015radiation} during the launch phase will constrain future deep space missions that might use nuclear power.

As with Space, in deep ocean environments, power constraints are also a current challenge. Current Autonomous Underwater Vehicles (AUVs) have limited capabilities due to limitations on energy storage and the availability of fuel sources. Furthermore, the extremely high pressures of deep-sea environments offer their challenges, not only to energy supply for AI but also to the mass and construction of protection containers for the electronics. Ocean glider AUV's use buoyancy engines with fins to convert force in the vertical direction to horizontal motion \cite{RN39,RN38,RN42,RN41,RN43}. While very slow, such UAV's are far less energy-constrained than other current technologies. However, the power generated by such engines is not currently suitable for powering AI systems. Batteries are used for such functions and must be recharged at the ocean surface using photo-voltaic cells. There are proposals to use nuclear fission power generation to enable deep-sea battery recharging stations for military UAV's, though these remain at the development stage and have similar safety considerations to those mentioned above for space \cite{RN59}.

\section*{Existence proof, human brains as efficient energy consumers}

The original goal of AI, which is still an important goal, was to extract principles from human intelligence. On the one hand, these principles would allow for a better understanding of intelligence and what makes us human. On the other hand, we could use those principles to build \emph{intelligent} artifacts, such as robots, computers, and machines. In both cases, the implication is to use human intelligence as a use case, which derives from the function of the brain.  We believe that there are also important energy efficiency principles that can be extracted from neurobiology and applied to AI. Therefore, nervous system can provide much inspiration for the construction of low power intelligent systems. 

The human nervous system is under tight metabolic constraints. These constraints include the essential role of glucose as fuel under conditions of non-starvation, the continuous demand for approximately 20 percent of the human body’s total energy utilization, and the lack of any effective energy-reserve among others~\cite{sokoloff1960metabolism}. And yet, as is well known, the brain operates on a mere 20 watts of power, approximately the same power required for the interior light of a typical household refrigerator. While being severely metabolically constrained is at one level a disadvantage, evolution has optimized brains in ways that lead to incredibly efficient representations of important environmental features that stand distinct from those employed in current digital computers.

The brain utilizes many means to reduce functional metabolic energy utilization. Indeed, one can observe at every level of the nervous system strategies to maintain high performance and information transfer, while minimizing energy expenditure. These range from ion channel distributions, to coding methods, to wiring diagrams (connectomes). Many of these methods could inspire new methodologies for constructing power efficient artificial intelligent systems.

At the neuronal coding level, the brain uses several strategies to reduce neural activity without sacrificing performance. Neural activity, (i.e., the generation of an action potential, the return to resting state, and synaptic processing) is very costly from an energetics standpoint, and this can determine the optimal number of spikes to encode either an engram or the neural representation of a new stimulus ~\cite{lennie2003cost,levybaxter96}. Such sparse coding strategies appear to be ubiquitous throughout the brain~\cite{beyeler2017sparse,olshausen1997sparse,olshausen2004sparse}.

Furthermore, dimensionality reduction methods from machine learning can explain many neural representations, especially in sensory systems~\cite{beyeler20163d}. Because brains face strict constraints on metabolic cost~\cite{lennie2003cost} and also consequent to the widespread existence of anatomical bottlenecks~\cite{ganguli2012compressed}, which often force the information stored in a large number of neurons to be compressed into an order of magnitude smaller population of downstream neurons, (for example, storing information from 100 million photoreceptors in 1 million optic nerve fibers), reducing the number of variables required to represent a particular stimulus space features prominently in efficient current coding theories of brain function~\cite{atick2011could, barlow2001redundancy, linsker1990perceptual}. Such views posit that the brain performs dimensionality reduction by maximizing mutual information between the high-dimensional input and the low-dimensional output of neuronal populations.  

The healthy brain must respond quickly to stimuli and changes in the environment. However, this implies an increase in neural activity, which would be energetically costly. Evidence suggests that the brain utilizes strategies to maintain a constant rate of activity. For example, the nervous system can respond quickly to perturbations by shifting the specific timing rather than increasing the absolute number of spikes~\cite{malyshev2013energy}. Moreover, the balance of excitation and inhibition can further maintain a steady rate of neural activity while still being responsive~\cite{sengupta2013balanced, yu2018efficient}. In these ways, the overall energy utilization of the human brain is relatively constant, while the local rate of energy consumption varies widely and is dependent upon functional neuronal activity and the balance between excitatory and inhibitory neurons ~\cite{Olds1994Sequential}  

At a macroscopic scale, the brain saves energy by minimizing the wiring between neurons and brain regions (i.e., number of axons), yet still communicates information at a high-level of performance~\cite{laughlin2003communication}. Unlike current electronic chips, the brain packs its wiring into a three-dimensional space, which not only reduces the overall volume but also can reduce the energy cost. Energy is further conserved by maintaining high local connectivity with sparse distal connectivity. White matter, which are myelinated axons that transmit information over long distances in the nervous system, make up about half the human brain but use less energy than gray matter (neuronal somata and dendrites) because of the scarcity of N+ and K+ ion channels along these axons~\cite{harris2012energetics}. These myelinated axons speed up signal propagation and reduce the volume of matter in the brain. However, information transfer between neurons and brain areas is still preserved by the overall architecture, which essentially is a small world network~\cite{sporns2006small,sporns2004small}. That is, even though the probability of any two distal cortical neurons being connected is extremely low, any two neurons are only a few connections away from each other. 
The nervous system also optimizes energy consumption at the cellular and sub-cellular levels. Glucose utilization reflects both the Na+-K+ ATPase used to repolarize axonal ionic gradients and the energy costs of neurotransmitter re-uptake. These together represent the majority of the brain’s energy budget and must be efficient for performance~\cite{engl2015non}. 


The brain's efficient power consumption may have a basis in thermodynamics and information theory. It has been suggested that any self-organizing system that is at equilibrium with its environment must minimize its free energy~\cite{friston2010free}. In other words, the system must adapt or evolve to resist a natural tendency toward disorder in an ever-changing environment. Top-down signals from downstream areas (e.g., frontal cortex or parietal cortex) can realize predictive coding~\cite{sengupta2013information,clark2013whatever, sengupta2013balanced}. Predicting outcomes reduces surprises, which leads to more energy efficient processing~\cite{friston2010free}. In this way organisms minimize the long-term average of surprise, which is the inverse of entropy, by predicting future outcomes. In essence, they minimize the expenditures required to deal with unanticipated events. The idea of minimizing free energy has close ties to many existing brain theories, such as the Bayesian brain, predictive coding, cell assemblies, and Infomax, as well as an evolutionary-inspired theory called Neural Darwinism or neuronal group selection~\cite{friston2010free}. 

The brain represents an important existence proof that extraordinarily efficient natural intelligence can compute in very complex ways within harsh, dynamic environments. Beyond an existence proof, brains provide an opportunity for reverse-engineering in the context of machine learning methods and neuromorphic computing.

\section*{Future AI} 
\subsection*{Edge computing in Remote and CMOS-hostile environments}

Unless human beings can be "radiation-hardened," robotic space probes will continue to dominate exploration and exploitation of space in domains ranging from low earth orbit to interstellar exploration. All of these domains are subject to a variety of hazards which are potentially hazardous to CMOS-based AI. These include collisions with high energy photons (such as gamma rays), micrometeorites, planetary weather, and anthropogenic attacks.

In many of these space domains, AI will be the preferred computational modality because of latency issues related to long-distance communication with Earth-based controllers. Operating in such domains will have the additional challenge of energetic constraints because readily available solar power may not always be available in domains such as the moon, solar system planets with weather and deep space (including interstellar). The primary alternative energy source for such domains is nuclear (both fission and fusion-based). Such power sources are in contrast to the current nuclear technologies used for missions such as the Mars Curiosity Rover. While break-even fusion power has yet to be demonstrated on Earth, the abundance of fusion fuels in the solar system makes such power sources attractive. In all these cases, the nuclear technology must have a similar resiliency to that of the AI in terms of hazards, and it will be optimal to consider such requirements holistically at the design stage.

Other such remote domains for AI exist. These include machines that must operate autonomously during the Antarctic winter, deep sea AI and military/law-enforcement applications. In each of these cases, the exact design criteria will be different, yet all will face unique power constraints that are in addition to the constraints to the AI itself.

\subsection*{Neuromorphic Computing and other Bio-Inspired Engineering Approaches} 

A key component in pursuing brain- and neural- inspired computing, coding, and algorithms lies in the currently shifting landscape of computing architectures. Moore's law, which has dictated the development of ever-smaller transistors since the 1960s, has become more and more difficult to follow, leading many to claim its demise~\cite{waldrop2016chips}.  This has inspired renewed interest in heterogeneous and non-Von Neumann computing platforms~\cite{chung2010single, shalf2015computing}, which take inspiration from the efficiency of the brain's architecture. Neuromorphic architectures can offer order-of-magnitude improvement in performance-per-Watt compared to traditional CPUs and GPUs~\cite{merolla2014million, indiveri2011neuromorphic, hasler2013finding}.  This enables, for example, IBM's TrueNorth chip to power convolutional neural networks for embedded gesture recognition at less than one Watt~\cite{amir2017low}.

Neuromorphic architectures refer to a wide variety of computing hardware platforms~\cite{schuman2017survey}, from sensing~\cite{posch2014retinomorphic, liu2010neuromorphic} to processing~\cite{merolla2014million, indiveri2011neuromorphic}, analog~\cite{fieres2008realizing} to digital~\cite{merolla2014million, furber2013overview}.  However, in most cases the defining characteristics take inspiration from the brain: 
\begin{enumerate} 
\item Massively parallel, simple integrating processing units (neurons)  
\item Sparse and dynamic low-precision communication via `spikes'.
\item Event-driven, asynchronous operation.
\end{enumerate}
This event-driven, distributed, processor-in-memory approach provides robust, low-power processing compatible with many neural-inspired machine learning and artificial intelligence applications~\cite{neftci2018data}.  Hence, size, weight and power (SWaP) constrained environments, such as edge and IoT devices, can leverage increased effective remote computation capabilities and provide real-time, low-latency intelligent and adaptive behavior. Moreover, the often noisy nature of learned artificial intelligence systems (some incorporate noise by design~\cite{srivastava2014dropout}) may lead to more robust computation in extreme environments such as space. 

\begin{figure}[!ht]
\centering
\includegraphics[width=1\linewidth]
{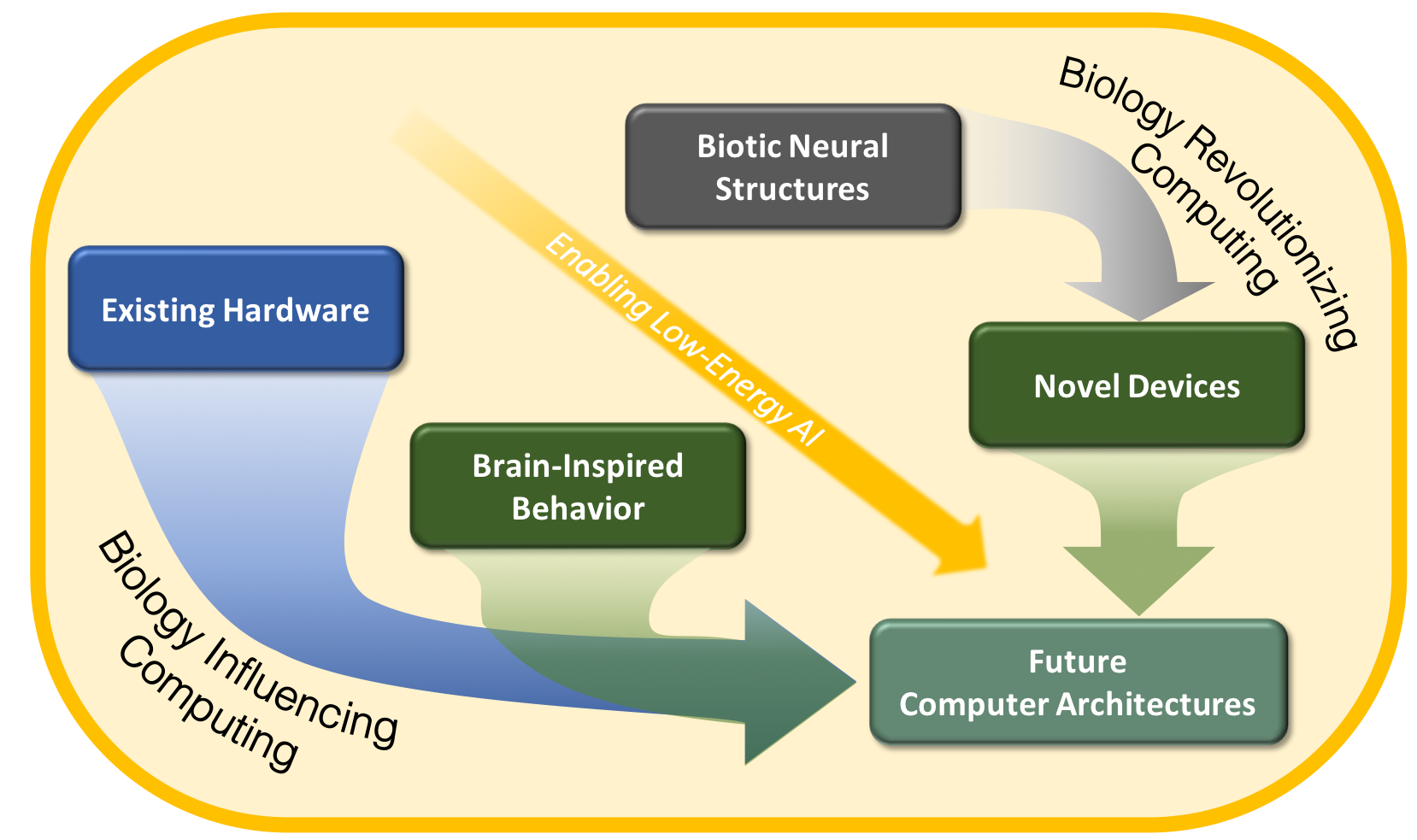}
\caption{The image shows that in the design of bio-mimetic circuitry, either the existing hardware will be slightly adjusted to copy the behavior of brain parts, or new computing architecture will be designed so that it completely emulate the high energy-efficient biotic neural structures (Image adapted from \cite{RN65}).}
\label{fig:neuromorphic}
\end{figure}

Power consumption in large-scale computing systems is also forcing the development of more efficient computing platforms~\cite{hemmert2010green}.  In data centers, heterogeneous (non-spiking) architectures are improving performance and latency, exemplified by a $30$--$80$x improvement on deep learning tasks~\cite{putnam2014reconfigurable,jouppi2017datacenter}, and new neuromorphic architectures, such as Intel's Loihi, are themselves heterogeneous which improves communication between neural and conventional cores~\cite{davies2018loihi}.  Emerging neural computing platforms may benefit traditional large-scale computation both indirectly (e.g., system health~\cite{das2018desh}, failure prediction~\cite{bouguerra2013failure}) and directly (e.g., meshing, surrogate models~\cite{melo2014development}), and recent work indicates that neuromorphic processors may be useful for direct computation due their high-communication, highly-parallel nature~\cite{aimone2017neural, severa2018spiking, jonke2016solving, lagorce2015stick}.

Neural inspiration has also impacted data collection in the form of spiking neuromorphic sensors which generally follow the same three characteristics as neuromorphic architectures. The two most common categories are silicon cochleas~\cite{chan2007aer, watts1992improved} and retina-inspired event-driven cameras~\cite{delbruck2014integration,delbruck2010activity}, though neuromorphic olfaction is also under active research and development~\cite{vanarse2016review}. Neuromorphic sensors can often be thought of as a method for high-speed preprocessing, fundamentally changing the sample space. For example, for imagery this allows far low-bandwidth, high-sampling, and high-dynamic range imagery~\cite{posch2015giving, delbruck2010activity}. These benefits, in turn, have enabled low-latency, low-power applications such as gesture recognition~\cite{amir2017low, ahn2011dynamic}, robotic control~\cite{delbruck2013robotic,conradt2009embedded}, and movement determination~\cite{haessig2018spiking, drazen2011toward}. Spiking sensors are innately compatible with spiking neuromorphic processors, and combining neuromorphic sensors with a neuromorphic processor can avoid the costly conversion between binary data formats and spikes.  

Computational requirements of artificial intelligence algorithms limit their remote applications today. Consequently, most current consumer or commercial machine learning technologies are reliant on connections to remote data centers.  However, as neuromorphic technologies transition from research platforms to everyday products, learning systems can and will proliferate in capability and scope. Combined with the expected growth of edge and IoT devices, we can expect persistence and pervasive learning devices. These learning devices, extensions of current trends in smart devices (e.g., digital assistants, smart home control, wearables), will be enhanced with personalized online learning and enabled with adaptive, intelligent and context-dependent perception and behaviors. Ultra-low energy neuromorphic chips will power remote computation at the milliwatt scale. In the industrial, medical and security spaces, the same technologies will provide low-powered sensors capable of extended deployment in a variety of extreme environments. Privacy, ethical, and environmental concerns will need to be balanced against convenience, productivity, and safety.

There is ample precedent for biologically-inspired approaches to engineering design that may lead to energy-efficient AI and edge computing. Both designing algorithms to mimic the brain's behavior, and building new computer hardware that mimic neural dynamics can lead to energy efficiency (see Figure \ref{fig:neuromorphic})\cite{RN65}. Furthermore, energy efficiency can be inspired by observing nature's non-neural solutions. For example, the wing of an airplane takes inspiration from the wings of flying animals (birds, bats, and insects). The shape of a modern naval submarine has evolved from early boat-like designs prevalent during the First and Second World Wars towards a more streamlined whale-like shape. Even DNA-based computation--by itself incredibly energetically efficient--takes inspiration from the conserved transcription information transfer mechanism of Earth's biosphere.

From a computational perspective, it is not only the brain which offers precedent. The adaptive immune system, with its sophisticated "learning and memory" through selection also represents a low-energy approach to artificial intelligence that may eventually have applications to AI-enhanced cyber-security applications~\cite{forbes2004imitation, forrest2007computer,somayaji1998principles, forrest1993genetic, rice2007using, keller2017leveraging}. This selectionist approach, which was inspired by the immune system, led to an influential brain theory where the synaptic selection took place during neural development and through experiential synaptic plasticity \cite{RN66,RN67}. Such a Darwinist approach can lead to efficient neural network structures. 






As edge computing and mobile sensing devices become ubiquitous, efficient mobility, whether on land, air, or water will become increasingly important. Biological organisms have evolved to leverage their environment, and this morphological computation can lead to efficient movement and information processing \cite{RN45}. For example, bipedal walking is somewhat of a controlled fall, where energy is conserved by allowing gravity to take over after the swing phase of a step. This strategy has been adopted in passive walker robots that utilize orders of magnitude less energy than conventional walking robots \cite{RN37, RN36}. Birds of prey and long-range migrating birds take advantage of thermal plumes to reduce energy usage during flight \cite{RN46, RN47,RN62}. Gliders have mimicked this strategy in their flight control systems \cite{RN49, RN48, RN34}. Similar to many fish and marine mammals, oceanographic submersible gliders can harvest energy from the heat flow of thermal gradients \cite{RN39,RN38,RN42,RN41,RN43}. These submersible gliders can operate across thousands of kilometers over months to years. Some fish species and flying insects alter their environment (i.e., the water or air vortices) to create additional thrust \cite{RN60,RN61}. Social insects and bacterial colonies have inspired highly distributed robots or computing systems \cite{RN63,RN64}. In these cases, each agent has very low power computation requirements, and no single agent is a point of failure. However, the interactions between these agents can lead to complex problem-solving. Taken together, future AI systems that take inspiration from biology and other energy harvesting approaches will have a distinct advantage for long-term operation in harsh or remote environments.

\begin{figure}[!ht]
\centering
\includegraphics[width=1\linewidth]
{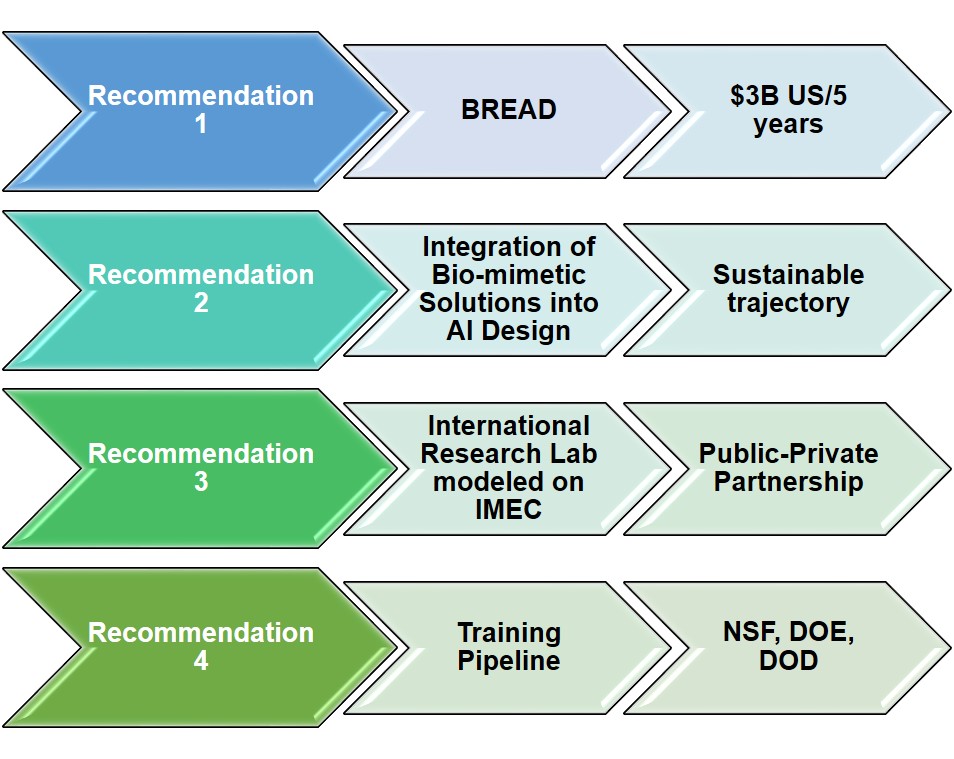}
\caption{A coherent strategy for ensuring that AI's of the future are not starved for power. The four recommendations presented in this Perspective work together to ensure government support for innovation, research, integrated design, and a human capital pipeline.}
\label{fig:network}
\end{figure}

\section*{Policy Recommendations}

AI is on a trajectory to fundamentally change society in much the same way that the industrial revolution did. Even without the development of General Artificial Intelligence, the trend is towards human-machine partnerships that collectively will have the ability to substantially extend the reach of humans in multiple domains (e.g., space, cyber, deep sea, nano). However, as with many things, there is no free lunch: AI will require energy inputs that we believe must be accounted for at all stages of the AI design process. We believe that such design solutions should leverage the solutions that biology, especially the human brain, has evolved to be energy efficient without sacrificing functionality. These solutions are critical components to what we call \emph{intelligence.}

\subsection*{Recommendation 1: A Multinational initiative to make BREAD} To accomplish the above, we advocate that under the auspices of the Association for Computing Machinery (ACM), American Psychological Association (APA),  Institute of Electrical and Electronics Engineers (IEEE), Society for Neuroscience, and multilateral governmental agreements, a global technological innovation initiative be launched, called Biomimetic Research for Energy-efficient, AI Designs (BREAD), to coordinate and catalyze research public and private investments into holistic AI design. Energy sustainability would be central in BREAD, but the initiative would encompass all aspects of AI from hardware to sensors.

Within the United States, we propose that President's Science Advisor should name the chair of a BREAD-inspired inter-agency coordinating sub-committee under the auspices of the National Science and Technology Council. In the US, BREAD would be truly cross-cutting and should receive substantial and explicit funding in relevant agency budgets including, DOE, NSF, DOD, and NIH. Given the stakes, an annual expenditure of approximately \$3B over five years would be appropriate. The United States portion of BREAD should aim to catalyze advances in bridging knowledge gleaned from the life sciences to AI energetics at US research institutions, the national labs and in coordination with industry. We additionally recommend that DOE stand-up a new BREAD office with overall government-wide budget authority analogous to NSF with STEM education.

\subsection*{Recommendation 2: Integrate biomimetic energetic solutions into future AI designs}  It is clear that AI will continue to be energetically constrained into the future. Even as this technological revolution continues to play out with ever more efficient strategies, thermodynamic considerations alone make the energetic considerations important. The source and availability of power to AI's will be critical to their function under all conditions, but particularly those at the Edge such as IoT, Space or other environments that are inherently hostile to CMOS (or future successor chip technologies). Further, because of the absolute requirement for adequate energy provision, future AI development will require an integrated design process where energy supply is not an add-on or assumed. Such integration of design already is prevalent in the design of new commercial aircraft and many medical devices; however, is has been hitherto remarkably absent in AI design. Indeed many instances of current AI (such as deep learning) approach the problem by purposeful situating of data centers close to sources of abundant and cheap electric power (such as hydro plants). We believe that a more fruitful approach should be to leverage the solutions evolved by biology (nervous system, metabolism, morphology, etc.) and embed such design thinking into future AI design. Thus, while deep learning AI's might continue to use the data center model for current and future applications that are similar to those used today by Google, Facebook, IBM, and other large corporations, AI operating in the context of Edge or otherwise constrained environments would use an architecture that integrate power provision and management based on biomimetic models.

\subsection*{Recommendation 3: A pre-competitive international research lab modeled on IMEC} As with CMOS prototype design and fabrication, the initial costs of integrating biomimetic solutions into AI are likely to be front-loaded. Hence, we recommend that stakeholder industrial partners come together with governments to build out a pre-competitive research laboratory modeled on IMEC in Leuven Belgium. IMEC, a public-private partnership, allows CMOS design firms across the globe to prototype new chips in a state-of-the-art environment that preserves intellectual property. In the context of BREAD, an IMEC model would serve a similar function for numerous companies active in the AI industrial space prior to production. Such a public-private partnership could catalyze the technological innovations necessary for success.

\subsection*{Recommendation 4: A Trainee Pipeline} We believe that it is critical to establish a pipeline of scientists who can integrate knowledge from biology, computer science, neuroscience, and engineering to build new human capital expertise at the intersection of AI and power management issues. To accomplish that, aligned with BREAD, research institutions should consider new graduate offerings at this nexus. In the United State, doctoral support for such students should be provided by the Engineering Directorate of the NSF and through the NSF's prestigious Graduate Research Fellowship Program. At the same time, there should be support for post-doctoral trainees in this area, both at the National Labs, but also through individual fellowships from either the Engineering Directorate of NSF, the Department of Energy or the Department of Defense.

\section*{Conclusion} In conclusion, we see the future development of AI as requiring new strategies for embedding energy demands of the machine into the overall design strategy. From our standpoint, this must include biomimetic solutions. As indicated above, there is much precedence for this type of engineering in other high aspects of technology, especially those that must operate in challenging environments. Now such engineering must be applied to future AI design so that the technological trajectory of this paradigm-changing technology is secure.

\bibliography{sample}




\section*{Acknowledgements}

This work was supported by USAF grant FA9550-18-1-0301 to JO. 

Sandia National Laboratories is a multi-mission laboratory managed and operated by National Technology and Engineering Solutions of Sandia, LLC, a wholly owned subsidiary of Honeywell International, Inc., for the U.S. Department of Energy's National Nuclear Security Administration under contract DE-NA0003525. This paper describes technical results and analysis.   Any subjective views or opinions that might be expressed in the paper do not necessarily represent the views of the U.S. Department of Energy or the United States Government.

\section*{Author contributions statement}

JO was PI on the USAF grant which supported this research. JK, WS, SK and JO wrote the manuscript.

\end{document}